\documentclass[12pt]{article}

\usepackage[a4paper, margin=1in]{geometry}
\usepackage{times}

\usepackage{graphicx}
\graphicspath{{./figures/}}

\usepackage{amsmath, amssymb, amsfonts}

\usepackage{setspace}
\usepackage{booktabs}
\usepackage{multirow}
\usepackage{makecell}
\usepackage{threeparttable}
\usepackage{tabularx}
\usepackage{adjustbox}

\usepackage{algorithm}
\usepackage{algorithmic}

\usepackage{float}
\usepackage{placeins}
\usepackage{subcaption}

\usepackage{tikz}
\usetikzlibrary{arrows.meta}
\usetikzlibrary{positioning, arrows.meta, calc}
\usepackage{pgfplots}
\usepgfplotslibrary{groupplots}
\pgfplotsset{compat=1.17}

\usepackage{xcolor}
\usepackage{hyperref}
\usepackage{soul}
\usepackage{textcomp}
\usepackage{orcidlink}
\usepackage{dashrule}

\begin{document}

\title{FAME: A Lightweight Spatio-Temporal Network for Model Attribution of Face-Swap Deepfakes}

\author{
  Wasim Ahmad\textsuperscript{1,2,3}\texttt{was\_last@iis.sinica.edu.tw}\thanks{Corresponding author. Email:
  \texttt{was\_last@iis.sinica.edu.tw}}
  \and
  Yan-Tsung Peng\textsuperscript{3} \texttt{ytpeng@cs.nccu.edu.tw} 
  \and
  Yuan-Hao Chang\textsuperscript{4}
  \texttt{johnson@csie.ntu.edu.tw}
}

\date{}
\maketitle

\begin{center}
\textsuperscript{1} Institute of Information Science, Academia Sinica, Taipei, Taiwan, 115 \\
\textsuperscript{2} Social Networks and Human-Centred Computing, Taiwan International Graduate Program, Taipei, Taiwan, 115 \\
\textsuperscript{3} Department of Computer Science, National Chengchi University, Taipei, Taiwan, 116 \\
\textsuperscript{4} Department of Computer Science and Information Engineering, National Taiwan University, Taipei, Taiwan, 106 \\
\end{center}

\vspace{1em}

\begin{center}
\textit{This is the accepted manuscript version of the article accepted in \textbf{Expert Systems with Applications}, June 2025. \\
Final version available at: \url{https://github.com/wasim004/FAME}}
\end{center}

\vspace{1em}

\begin{abstract}
The widespread emergence of face-swap Deepfake videos poses growing risks to digital security, privacy, and media integrity, necessitating effective forensic tools for identifying the source of such manipulations. Although most prior research has focused primarily on binary Deepfake detection, the task of model attribution—determining which generative model produced a given Deepfake—remains underexplored. In this paper, we introduce \textbf{FAME} (Fake Attribution via Multilevel Embeddings), a lightweight and efficient spatio-temporal framework designed to capture subtle generative artifacts specific to different face-swap models. FAME integrates spatial and temporal attention mechanisms to improve attribution accuracy while remaining computationally efficient. We evaluate our model on three challenging and diverse datasets: Deepfake Detection and Manipulation (DFDM), FaceForensics++ (FF++), and FakeAVCeleb (FAVCeleb). Results show that FAME consistently outperforms existing methods in both accuracy and runtime, highlighting its potential for deployment in real-world forensic and information security applications. Code and pretrained models: \url{https://github.com/wasim004/FAME/}.
\end{abstract}

\vspace{0.5em}

\noindent\textbf{Keywords:} Face-swap Deepfakes, Deepfake Model Attribution, Attention Mechanism, Multimedia Forensics, Information Security

\section{Introduction}
\label{sec:intro}
The term \textit{Deepfake} refers to synthetic media generated using deep learning, most commonly involving the realistic swapping of one individual's face onto another in video content. With the rise of open-source tools such as DeepFaceLab~\cite{deepFaceLab} and FaceSwap~\cite{faceswap}, generating Deepfake videos has become increasingly accessible~\cite{chesney2019deep}. While these tools offer legitimate applications in entertainment and accessibility, they have also been misused for identity fraud, misinformation, and political manipulation~\cite{9105991}, raising serious concerns for privacy, security, and digital trust.

In response, numerous Deepfake detection techniques have emerged, including methods based on visual inconsistencies~\cite{li2018exposing, mittal2020emotions, guarnera2020deepfake}, frequency domain analysis~\cite{wang2022forgerynir}, and deep neural networks~\cite{afchar2018mesonet, de2020deepfake, kim2021fretal, 8682602}. These advances have been fueled by benchmark datasets such as FaceForensics++ (FF++)~\cite{rossler2019faceforensics++}, Celeb-DF~\cite{li2020celeb}, DFDC~\cite{dolhansky2019deepfake}, DeeperForensics~\cite{jiang2020deeperforensics}, and WildDeepfake~\cite{zi2020wilddeepfake}, which capture diverse manipulation techniques, including expression reenactment~\cite{song2021pareidolia, burkov2020neural}, face swapping~\cite{zhu2021one, agarwal2021detecting}, and attribute editing~\cite{xu2021facecontroller, fu2021high}.

While binary detection (real vs. fake) remains foundational, a critical yet underexplored task in multimedia forensics is \textit{model attribution}—identifying the specific generative model or tool used to create a Deepfake. Attribution enables traceability and supports investigations by narrowing down the techniques or actors involved. Although some recent studies have attempted GAN-based attribution~\cite{girish2021towards}, such methods are less effective for face-swap Deepfakes, where encoder-decoder pipelines tend to obscure high-frequency generative artifacts~\cite{dolhansky2020deepfake}. Face-swap Deepfakes present a distinct forensic challenge: their outputs often appear visually similar across models, yet they encode subtle decoder-specific artifacts. Accurate attribution requires capturing both fine-grained spatial cues and temporal dynamics—features often overlooked by global classification models or overly complex Transformer-based systems that are impractical in forensic settings.
\newline\newline
In this work, we introduce \textbf{FAME} (Fake Attribution via Multi-level Embeddings), a lightweight and domain-specific spatio-temporal framework for fine-grained model attribution of face-swap Deepfakes. FAME is designed to uncover generative signatures embedded in video content by leveraging attention-based mechanisms that highlight subtle but consistent traces left by different synthesis pipelines. It is compact (2.61M parameters), efficient, and robust across diverse manipulation types and compression settings.

To validate our approach, we evaluate FAME on three challenging datasets: DFDM, which features face-swap videos generated using encoder-decoder variants; FF++, a widely used benchmark with diverse manipulation techniques; and FakeAVCeleb~\cite{khalid2021fakeavceleb}, which introduces multimodal (audio-visual) Deepfakes. Each dataset poses a unique attribution challenge, allowing us to test the generalizability and effectiveness of the proposed framework.
\newline\newline
\textbf{While FAME utilizes well-known components such as VGG-19, LSTM, and attention mechanisms, \textit{its novelty lies in their task-specific integration for the underexplored domain of model attribution} in face-swap Deepfakes.} Unlike binary detection tasks that classify content as real or fake, model attribution demands differentiation among highly similar outputs generated by distinct synthesis pipelines. FAME introduces a multi-level attention strategy optimized to capture fine-grained decoder-specific traces, and employs a hybrid spatial-temporal loss formulation. These design decisions are not only computationally efficient but specifically tailored to highlight generative artifacts, enabling robust attribution across diverse datasets.

Our main contributions are as follows:

\begin{itemize}
    \item We propose \textbf{FAME}, a lightweight spatio-temporal framework tailored for the attribution of face-swap Deepfake models based on fine-grained generative cues.
    \item We conduct a comprehensive evaluation across three diverse datasets—DFDM, FF++, and FakeAVCeleb—demonstrating state-of-the-art attribution performance and strong cross-dataset generalization.
    \item We provide detailed runtime analysis and benchmarking against existing methods, showing that FAME is both accurate and computationally efficient for forensic applications.
\end{itemize}

\section{Related Work}

\subsection{Deepfake Detection}
Deepfake detection research has focused primarily on distinguishing between real and manipulated content. Several benchmark datasets have supported this effort, including FaceForensics++ (FF++)~\cite{rossler2019faceforensics++}, Celeb-DF~\cite{li2020celeb}, and DFDC~\cite{dolhansky2019deepfake}, which contain large-scale collections of real and Deepfake videos. However, these datasets are primarily geared toward binary classification tasks and often lack detailed annotations about the generative models used, making them less suitable for model attribution.

Detection techniques typically fall into two categories: artifact-based and learning-based. Artifact-based methods identify telltale inconsistencies in head pose~\cite{chesney2019deep}, emotion~\cite{de2020deepfake}, frequency spectra~\cite{durall2019unmasking}, or video signal patterns~\cite{li2018exposing, dolhansky2020deepfake}. Learning-based methods leverage deep neural networks, including capsule networks~\cite{8682602}, ensemble CNNs~\cite{bonettini2021video}, attention models~\cite{zhao2021multi}, and spatio-temporal architectures~\cite{gu2021spatiotemporal, kim2021fretal}. More recently, vision Transformers such as ViViT~\cite{arnab2021vivit}, TimeSformer~\cite{bertasius2021space}, ResVit~\cite{ahmad2022resvit}, and VideoSwin~\cite{liu2022videoswin} have gained traction due to their ability to capture global and temporal relationships however, their high parameter counts (often >100M) and large compute requirements make them impractical for forensic deployments yet focus on classification, not attribution. Several hybrid and multi-modal frameworks have also emerged e-g. Hashmi et al.~\cite{9980255} proposed an audiovisual ensemble for FakeAVCeleb. However, these approaches excel in detecting manipulation, they are not designed to determine \textit{which} model created the Deepfake—a critical requirement in forensic investigations. Artifact-based cues may generalize across Deepfakes regardless of their origin, offering limited attribution capability. In contrast, the proposed \textbf{FAME} framework prioritizes attribution accuracy while maintaining efficiency. 

Model attribution requires the extraction of subtle, decoder-specific cues that may be lost in models optimized solely for detecting general inconsistencies or high-frequency noise. FAME addresses this by focusing on decoder-level differentiation through spatio-temporal refinement and dataset-wide generalization.

\subsection{Model Attribution}
Model attribution aims to identify the specific generative model responsible for creating a given synthetic image or video~\cite{karras2020analyzing}. This problem has gained attention in the context of GAN-generated content, where researchers have explored fingerprinting techniques~\cite{asnani2023reverse, mittal2020emotions}, frequency-based signatures~\cite{durall2019unmasking}, and source classification~\cite{jiang2020deeperforensics}. However, these works largely focus on still images or GAN-based Deepfakes, and are not directly applicable to face-swap videos generated via autoencoder-based pipelines.

Face-swap Deepfakes pose a unique challenge for attribution due to their subtle, decoder-specific visual artifacts. Unlike GANs, deepfake autoencoders (DFAEs) tend to smooth high-frequency details~\cite{dolhansky2020deepfake}, making source differentiation more difficult. Despite this, recent work has shown that it is possible to extract discriminative features from DFAE-generated content.

CapST~\cite{Ahmad2025CapST} is one of the few models that target attribution in this space, combining capsule networks with temporal attention for video attribution based on DFAE. Building on this idea, we propose \textbf{FAME}, a lightweight spatio-temporal attention-based model tailored for fine-grained attribution of face-swap Deepfakes. Unlike CapST, FAME adopts a simplified design optimized for efficiency, yet delivers superior performance across three diverse datasets—DFDM, FF++, and FakeAVCeleb.

In contrast to prior work that either targets binary detection or focuses narrowly on GAN attribution, FAME addresses the broader challenge of efficient model attribution for face-swap Deepfakes. Its domain-specific design and cross-dataset generalization make it a strong candidate for real-world forensic deployment. With just 2.61M parameters, FAME captures discriminative spatial-temporal patterns without relying on large-scale attention backbones—making it well-suited for constrained environments. While integrating efficient Transformer modules remains a compelling avenue for future work, our focus is on model-level attribution rather than binary detection.

\section{Methodology}

\subsection{Datasets: DFDM, FF++, and FAVCeleb}

The DFDM dataset serves as a cornerstone for Deepfake model attribution research due to its unique focus on labeled face-swap Deepfakes generated by multiple Autoencoder-based architectures. It includes videos created using tools such as FaceSwap and DeepFaceLab, specifically concentrating on five models: Faceswap (baseline), Lightweight, IAE, Dfaker, and DFL-H128. These models were selected based on subtle architectural variations that introduce distinct generative artifacts helpful for attribution~\cite{faceswap, deepFaceLab, dfaker, perov2020deepfacelab}. Real videos from the Celeb-DF dataset were used as source material, with face regions extracted using the S3FD detector and aligned via the FAN face aligner~\cite{li2020celeb, zhang2017s3fd, bulat2017far}. Each model was trained for 100,000 iterations, and the resulting Deepfakes were encoded in MPEG4.0 format under three H.264 compression levels: lossless, high quality, and low quality. The dataset contains a total of 6,450 Deepfake videos. A summary of the encoder-decoder architectures of the DFDM generation models is provided in Table~\ref{tab:deepfake_models}.

\begin{table}[ht]
  \caption{Architectural Settings of Deepfake Generation Models~\cite{jia2022model}}
  \label{tab:deepfake_models}
  \small
  \centering
  \begin{tabular}{@{}lccc@{}}
    \toprule
    Model & Input & Output & Encoder \& Decoder Design \\ 
    \midrule
    Faceswap & 64 & 64 & 4 Convs + 3 Upsamples + 1 Conv \\ 
    (baseline) & & & \\
    Lightweight & 64 & 64 & 3 Convs + 3 Upsamples + 1 Conv \\
    IAE & 64 & 64 & 4 Convs + 4 Upsamples + 1 Conv \\
    \multirow{2}{*}{Dfaker} & \multirow{2}{*}{64} & \multirow{2}{*}{128} & 4 Convs + 4 Upsamples \\ 
    & & & 3 Residuals + 1 Conv \\
    DFL-H128 & 128 & 128 & 4 Convs + 3 Upsamples + 1 Conv \\
    \bottomrule
  \end{tabular}
  \begin{tablenotes}
    \scriptsize
    \item ‘Convs’ and ‘Upsamples’ denote convolutional and upsampling layers. In IAE, the encoder and decoder share intermediate layers.
  \end{tablenotes}
\end{table}

To ensure broader evaluation and generalizability, we also utilize the FaceForensics++ (FF++) and FakeAVCeleb (FAVCeleb) datasets. FF++ is widely used in the Deepfake detection community due to its diverse manipulation techniques, including face reenactment and identity swapping. It features multiple manipulation types under varying compression levels, which introduce a range of synthetic artifacts. The architectural settings of the generation methods used in FF++ are summarized in Table~\ref{tab:ffpp}.

\begin{table}[ht]
\centering
\caption{Architectural Settings of FaceForensics++ Dataset Manipulation Models~\cite{rossler2019faceforensics++}}
\resizebox{\linewidth}{!}{
\begin{tabular}{@{}llll@{}}
\toprule
\textbf{Model}         & \textbf{Input}   & \textbf{Output}  & \textbf{Encoder \& Decoder Design}                                 \\ \midrule
Deepfakes (DF)         & 64 × 64          & 64 × 64          & Encoder-Decoder with 4 Convs, 3 Upsamples, and 1 Conv     \\
Face2Face (F2F)        & Variable         & Variable         & Traditional 3D facial reconstruction pipeline                      \\
FaceSwap (FS)          & Variable         & Variable         & 3D reconstruction-based identity-swapping pipeline               \\
NeuralTextures (NT)    & 128 × 128        & 128 × 128        & Neural texture synthesis with 3D model-based rendering          \\ \bottomrule
\end{tabular}
}
\label{tab:ffpp}
\end{table}

For FakeAVCeleb, we selectively include only five of its seven available classes, focusing exclusively on video-based Deepfakes relevant to our visual-only attribution task. The dataset is known for high-quality multi-modal Deepfakes, but we limit our use to the visual modality to maintain task alignment. The included techniques vary from standard face-swap models to more advanced GAN-based systems, which introduce synchronization artifacts and lip movement subtleties. Their architectural descriptions are presented in Table~\ref{tab:fakeavceleb}.

\begin{table}[ht]
\centering
\caption{Architectural Settings of FakeAVCeleb Dataset Manipulation Models~\cite{khalid2021fakeavceleb}}
\resizebox{\linewidth}{!}{
\begin{tabular}{@{}llll@{}}
\toprule
\textbf{Model/Technique}       & \textbf{Input}   & \textbf{Output}  & \textbf{Architecture/Design}                                      \\ \midrule
FaceSwap                      & Variable         & Variable         & Encoder-decoder ConvNet for face-swapping                         \\
FSGAN                         & Variable         & Variable         & GAN-based system with face alignment and reenactment              \\
SV2TTS (Audio Cloning)        & Variable         & Variable         & Three-stage pipeline: speaker encoder, synthesizer, vocoder       \\
Wav2Lip (Lip Sync)            & Variable         & Variable         & GAN-based architecture for lip synchronization                    \\ \bottomrule
\end{tabular}
}
\label{tab:fakeavceleb}
\end{table}

These architectural distinctions are critical, as model attribution relies on detecting decoder-specific traces that emerge from the internal structure of these generation pipelines. Together, these datasets form a robust foundation for evaluating the effectiveness and generalization capability of the proposed framework. Finally, Table~\ref{tab:dataset_stats} summarizes the core characteristics of each dataset.

\begin{table}[ht]
\centering
\caption{Dataset Characteristics and Statistics}
\resizebox{\linewidth}{!}{%
\begin{tabular}{l|c|c|c}
\hline
\textbf{Property}                    & \textbf{DFDM} & \textbf{FF++} & \textbf{FAVCeleb} \\
\hline
Number of Videos (Total)            & 6,450         & $\sim$1,000+  & $\sim$2,000+      \\
Real / Fake Ratio                   & 1:5           & 1:1           & 1:1               \\
Number of Deepfake Models           & 5             & 4             & 5                 \\
Resolution                          & $112 \times 112$ (preprocessed) & Varies (up to $128 \times 128$) & Variable (High-Quality) \\
Average Video Duration              & 5–10 seconds  & 10–20 seconds & 5–15 seconds      \\
Number of Frames Extracted          & 10 per video  & 10 per video  & 10 per video      \\
Face Alignment                      & OpenFace + FAN & FF++-specific & OpenFace + FAN    \\
Train / Test Split                  & 80 / 20       & 70 / 30       & 70 / 30           \\
Compression Levels                  & None, HQ, LQ  & HQ, LQ        & HQ only           \\
\hline
\end{tabular}
}
\label{tab:dataset_stats}
\end{table}

\begin{table}[H]
\centering
\caption{FAME Model Architecture and Training Hyperparameters}
\begin{tabular}{l|l}
\hline
\textbf{Parameter}           & \textbf{Value} \\
\hline
Input Size                  & $112 \times 112 \times 3$ \\
Base Feature Extractor     & VGG-19 (layers 0–26, BatchNorm included) \\
Activation Function        & ReLU \\
Batch Size                 & 32 \\
Number of Epochs           & 150 \\
Learning Rate              & 0.01, decayed $\times 0.1$ every 40 epochs \\
Optimizer                  & AdamW \\
Weight Decay               & 0.6 \\
Loss Function              & Weighted Cross-Entropy (Spatial + Temporal) \\
Temporal Aggregation       & LSTM with Attention \\
Dropout                    & Not used \\
Data Augmentation          & Random horizontal flip, resizing, normalization \\
Temporal Clip Sampling     & Random temporal crop with stride sampling \\
\hline
\end{tabular}
\label{tab:FAME-params}
\end{table}

\begin{figure}[t]
  \centering
  \includegraphics[height=7cm]{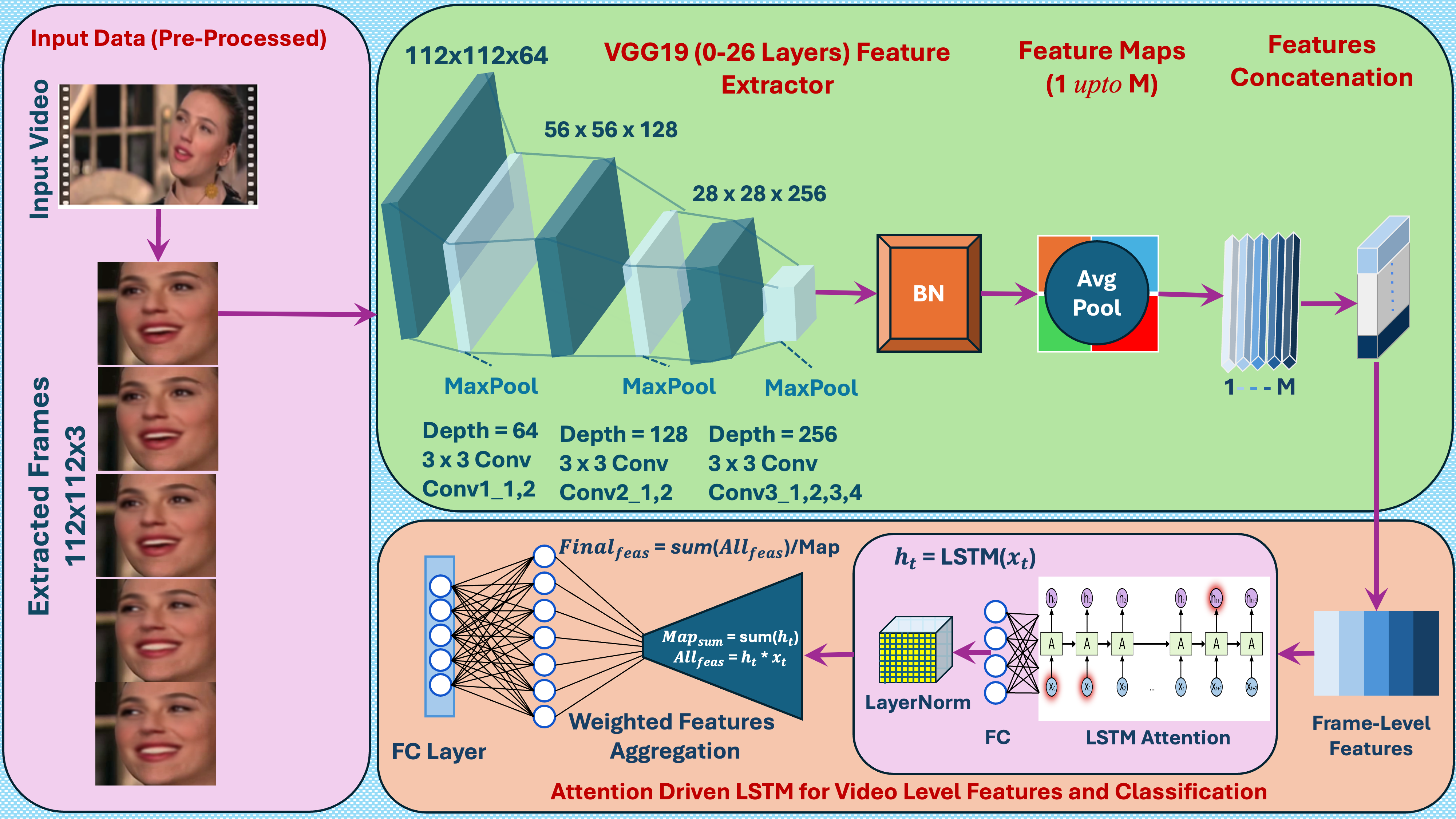}
  \caption{Architecture of the proposed Fine-Grained Attribution via Multi-level Attention (FAME) for Deepfake model attribution. The framework processes extracted face frames using a truncated VGG-19 network to obtain spatial features. These are then passed through an attention-enhanced bidirectional LSTM, which computes frame-level attention weights and aggregates temporal features. The final video-level representation is obtained via weighted feature aggregation and passed through a fully connected layer for model attribution. Both spatial and temporal attention mechanisms allow FAME to focus on subtle decoder-specific artifacts across frames.}
  \label{fig:main_arch}
\end{figure}


\begin{algorithm}
\small
\setstretch{0.9} 
\caption{FAME: Feature Attribution via Multilevel Embeddings for Deepfake Model Attribution}
\label{alg:vgg_lstm_algo}
\begin{algorithmic}[1]
\REQUIRE Video sample $X \in \mathbb{R}^{C \times T \times H \times W}$, with $T$ frames
\ENSURE Predicted class label $\hat{y} \in \mathbb{R}^{K}$ for $K$ deepfake generation models

\STATE \textbf{1. Spatial Feature Extraction:}
\FOR{each frame $X_t \in \mathbb{R}^{C \times H \times W}, \, t = 1 \dots T$}
    \STATE Extract spatial features via pretrained VGG-19:
    \[
    F_t = \phi_{\operatorname{VGG}}(X_t)
    \]
    \STATE Apply batch normalization and ReLU activation:
    \[
    \widetilde{F}_t = \operatorname{AvgPool}\left( \operatorname{BN}\left( \operatorname{ReLU}(F_t) \right) \right) \in \mathbb{R}^{D}
    \]
\ENDFOR
\STATE Stack frame features into matrix:
\[
R = [\widetilde{F}_1^\top, \dots, \widetilde{F}_T^\top]^\top \in \mathbb{R}^{T \times D}
\]
\STATE \textbf{2. Temporal Encoding with Attention:}
\STATE Encode temporal dynamics using a Bi-LSTM:
\[
H = \operatorname{BiLSTM}(R) \in \mathbb{R}^{T \times H_d}
\]
\STATE Generate frame-wise attention weights:
\[
A = \sigma\left( \operatorname{LayerNorm}(W_a H + b_a) \right) \in \mathbb{R}^{T \times D}
\]
\STATE Apply attention to spatial features:
\[
\widetilde{R}_t = A_t \odot R_t, \quad \forall t = 1, \dots, T
\]
\STATE \textbf{3. Temporal Aggregation and Classification:}
\STATE Compute the weighted global representation:
\[
z = \frac{1}{\sum_{t=1}^T A_t} \sum_{t=1}^T \widetilde{R}_t
\]
\STATE Apply dropout and project to class logits:
\[
\hat{y} = \operatorname{FC}(\operatorname{Dropout}(z))
\]
\RETURN $\hat{y}$
\end{algorithmic}
\end{algorithm}

\subsection{Fine-Grained Attribution via Multi-level Attention Attention (FAME) Architecture}

This study presents a novel framework, FAME, specifically designed to enhance the attribution of face-swap Deepfake videos by leveraging both spatial and temporal attention mechanisms. The architecture integrates a truncated VGG-19 network for spatial feature extraction with a bidirectional LSTM module, augmented by attention layers to capture temporal dependencies across frames. These features are subsequently aggregated and passed through a fully connected layer for final classification. Although FAME builds upon established deep learning components, its tailored configuration, combining spatial attention from VGG features, temporal attention via LSTM, and a domain-specific hybrid loss function, is uniquely optimized for the model attribution task. This contrasts with previous attention-based Deepfake detection approaches, which primarily aim to identify manipulated content without discerning its generative origin. The complete architectural design and processing pipeline are depicted in Figure~\ref{fig:main_arch} and detailed in Algorithm~\ref{alg:vgg_lstm_algo}, while the training and model configuration settings are summarized in Table~\ref{tab:FAME-params}.

\subsubsection{Frame-Level Feature Extraction}

The selection of layers 0–26 from VGG-19 ensures that mid-level spatial features—critical for identifying subtle decoder-specific artifacts—are retained without incurring excessive computational cost. This truncation balances semantic depth and spatial detail, making it suitable for forensic tasks like model attribution, where subtle differences are key. Given a sequence of input frames \( \{X_j\}_{j=1}^{M} \), where \( M \) is the number of frames per video clip, convolutional features are computed as:

\begin{equation}
F_j = \text{VGG}(X_j)
\end{equation}

These features are then processed through a ReLU activation function, batch normalization, and global average pooling to reduce dimensionality and enhance stability:

\begin{equation}
R_j = \text{AvgPool}(\text{BatchNorm}(\text{ReLU}(F_j)))
\end{equation}

The resulting vectors \( \{R_j\} \) represent each frame’s condensed spatial information and are passed into the temporal modeling stage. We further refine both the spatial and temporal features using attention modules, which are detailed in Section~\ref{subsec:attn_modules}.

\subsubsection{Attention Modules}
\label{subsec:attn_modules}

To improve fine-grained attribution performance, FAME integrates both spatial and temporal attention mechanisms. These help the model focus on decoder-specific artifacts in both individual frames and their temporal evolution.

\begin{figure}[ht]
\centering
\resizebox{\linewidth}{!}{ 
\begin{tikzpicture}[
    node distance=1.1cm and 1.2cm,
    every node/.style={font=\small},
    box/.style={draw, rectangle, minimum width=2.1cm, minimum height=0.9cm, align=center},
    arrow/.style={-Stealth, thick}
]


\node[box] (feat) {\( \mathbf{F} \in \mathbb{R}^{C \times H \times W} \)};
\node[box, below left=1.1cm and 1.4cm of feat] (avgpool) {Global \\ AvgPool};
\node[box, below right=1.1cm and 1.4cm of feat] (maxpool) {Global \\ MaxPool};
\node[box, below=2cm of feat] (mlp) {Shared MLP \\ + Sigmoid};
\node[box, below=1.3cm of mlp] (mask) {\( \mathbf{M}_s \in \mathbb{R}^{1 \times H \times W} \)};
\node[box, below=1.3cm of mask] (out) {\( \mathbf{F}' = \mathbf{F} \odot \mathbf{M}_s \)};

\draw[arrow] (feat) -- (avgpool);
\draw[arrow] (feat) -- (maxpool);
\draw[arrow] (avgpool) -- (mlp);
\draw[arrow] (maxpool) -- (mlp);
\draw[arrow] (mlp) -- (mask);
\draw[arrow] (mask) -- (out);
\draw[arrow] (feat.south) to[out=-90, in=180] (out.west);

\node at ($(feat.north)+(0,0.7)$) {\textbf{(a) Spatial Attention Module}};


\node[box, right=6.8cm of feat] (h1) {\( \mathbf{h}_1 \)};
\node[box, right=of h1] (h2) {\( \mathbf{h}_2 \)};
\node[box, right=of h2] (dots) {\( \cdots \)};
\node[box, right=of dots] (hT) {\( \mathbf{h}_T \)};

\node[box, below=1.1cm of h1] (e1) {\( e_1 \)};
\node[box, below=1.1cm of h2] (e2) {\( e_2 \)};
\node[box, below=1.1cm of hT] (eT) {\( e_T \)};

\node[box, below=1.1cm of e1] (a1) {\( \alpha_1 \)};
\node[box, below=1.1cm of e2] (a2) {\( \alpha_2 \)};
\node[box, below=1.1cm of eT] (aT) {\( \alpha_T \)};

\node[box, below=1.6cm of a2, minimum width=2.2cm] (z) {\( \mathbf{z} = \sum \alpha_t \cdot \mathbf{h}_t \)};

\draw[arrow] (h1) -- (e1);
\draw[arrow] (h2) -- (e2);
\draw[arrow] (hT) -- (eT);

\draw[arrow] (e1) -- (a1) node[midway, left] {\footnotesize softmax};
\draw[arrow] (e2) -- (a2);
\draw[arrow] (eT) -- (aT);

\draw[arrow] (h1.south) to[out=-140,in=160,looseness=2.5] (z.north west);
\draw[arrow] (a1.south) to[out=-90,in=150,looseness=1.3] (z.north west);

\draw[arrow] (h2.south) to[out=-135,in=122,looseness=2.2] (z.north);
\draw[arrow] (a2.south) to[out=-80,in=75,looseness=2.0] (z.north);

\draw[arrow] (hT.south) to[out=-140,in=40,looseness=2.5] (z.north);
\draw[arrow] (aT.south) to[out=-90,in=20,looseness=1] (z.north);

\node at ($(h2.north)+(0,0.7)$) {\textbf{(b) Temporal Attention Module}};

\end{tikzpicture}
} 
\caption{Illustration of attention mechanisms in FAME. (a) The spatial attention module computes a mask \( \mathbf{M}_s \) from pooled convolutional features and applies it to emphasize key regions in each frame. (b) The temporal attention module assigns weights \( \alpha_t \) to LSTM outputs \( \mathbf{h}_t \), enabling the model to focus on the most informative frames during sequence-level aggregation.}
\label{fig:attention_modules}
\end{figure}
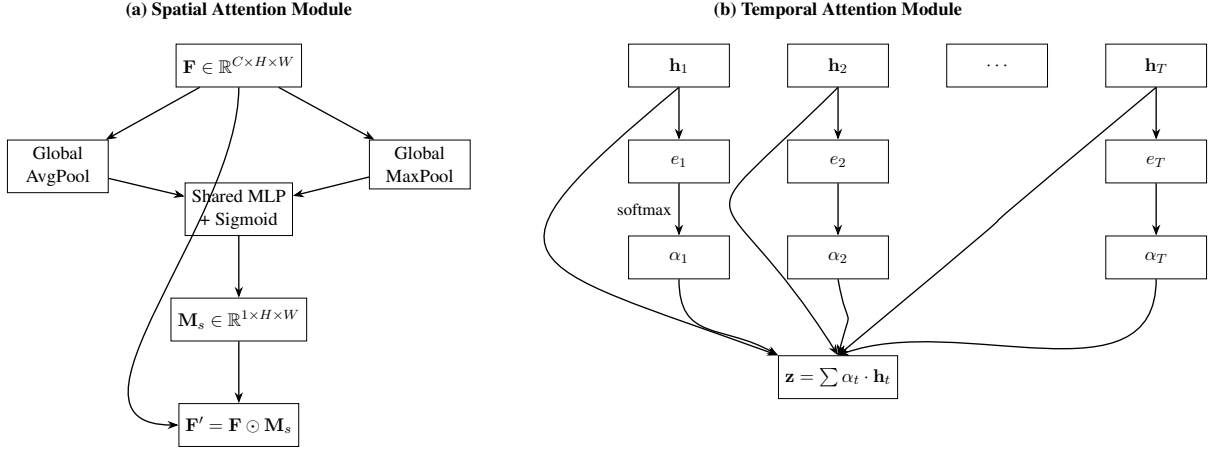

\textbf{Spatial Attention Module:} The spatial attention block operates on convolutional features extracted from each frame. Given a feature map \( \mathbf{F} \in \mathbb{R}^{C \times H \times W} \), we apply global average pooling and global max pooling across the channel dimension to summarize spatial activations. These pooled features are passed through a shared multi-layer perceptron (MLP) followed by a sigmoid activation to compute a spatial attention mask \( \mathbf{M}_s \in \mathbb{R}^{1 \times H \times W} \). The refined spatial feature map is obtained via element-wise multiplication as shown in Figure~\ref{fig:attention_modules}(a).

\begin{equation}
\mathbf{F}' = \mathbf{F} \odot \mathbf{M}_s
\end{equation}

This process emphasizes regions that contain salient generative artifacts.
\newline

\textbf{Temporal Attention Module:} Figure~\ref{fig:attention_modules}(b) illustrates the attention-refined frame embeddings \( \{\mathbf{F}'_t\}_{t=1}^T \), which are sequentially passed through a bidirectional LSTM to capture temporal dependencies, resulting in hidden states \( \{\mathbf{h}_t\} \). We adopt an attention-enhanced bidirectional LSTM instead of transformer-based encoders due to its lower parameter count and higher efficiency on short video sequences. This makes it more suitable for real-time or resource-constrained forensic applications, while still effectively modeling temporal relationships. To determine the relative importance of each frame in the context of attribution, we compute attention scores:

\begin{equation}
e_t = \mathbf{v}^T \tanh(\mathbf{W}_h \mathbf{h}_t + \mathbf{b})
\end{equation}

These scores are normalized using softmax to produce temporal attention weights \( \alpha_t \):

\begin{equation}
\alpha_t = \frac{\exp(e_t)}{\sum_{k=1}^{T} \exp(e_k)}
\end{equation}

The final clip-level representation is computed as a weighted sum of the hidden states:

\begin{equation}
\mathbf{z} = \sum_{t=1}^{T} \alpha_t \cdot \mathbf{h}_t
\end{equation}

This enables the model to emphasize the most informative frames for model attribution. A visual summary of both modules is provided in Figure~\ref{fig:attention_modules}.


\subsubsection{Temporal Feature Aggregation}
The frame-level features \( \{R_j\} \) are processed by an LSTM with attention mechanisms. The LSTM generates attention weights, and the temporal attention map is given by:
\begin{equation}
\text{Attn:Map} = \text{LSTM}(R)
\end{equation}
The aggregated feature is then computed as \( R^{ta} = \text{Attn:Map} \odot R \).


\subsubsection{Feature Aggregation and Classification}
The attention-weighted features are summed and normalized to obtain the final feature vector \( F_f = \frac{\sum_{j=1}^{M} (R^{ta})_j}{\sum_{j=1}^{M} \text{Attn:Map}_j} \). This vector is then passed through a fully connected layer to produce the predicted probabilities for different Deepfake models \( \hat{y} = \text{fc}(F_f) \).

Our loss function integrates spatial and temporal features to improve model attribution, ensuring that the model captures information at the frame-level and the sequence-level. The loss of spatial characteristics, \( \mathcal{L}_{\text{spatial}} \), is calculated using cross-entropy loss at the frame level:

\begin{equation}
\mathcal{L}_{\text{spatial}} = -\sum_{i=1}^{N} y_i \log(\hat{y}_i^{\text{spatial}})
\end{equation}

where \( y_i \) is the true label and \( \hat{y}_i^{\text{spatial}} \) represents the predicted probability of the spatial features.

The loss for temporal features is similarly defined using cross-entropy at the sequence level:

\begin{equation}
\mathcal{L}_{\text{temporal}} = - \sum_{j=1}^{M} y_j \log(\hat{y}_j^{\text{temporal}})
\label{eq:temporal_loss}
\end{equation}

where \( y_j \) is the true label for the \( j \)-th sequence.

The total loss is a weighted combination of the spatial and temporal components:

\begin{equation}
\mathcal{L}_{\text{total}} = \alpha \mathcal{L}_{\text{spatial}} + \beta \mathcal{L}_{\text{temporal}}
\end{equation}

where \( \alpha \) and \( \beta \) are the weights assigned to the spatial and temporal loss terms, respectively. This formulation ensures that FAME learns discriminative patterns from both the spatial domain and the sequential frame dynamics and thus improves the attribution of the Deepfake model by focusing on the most informative aspects of the data.

Although the FAME architecture draws from well-established CNN and LSTM structures, its contribution lies in the domain-specific configuration tailored for Deepfake model attribution. The proposed spatial-temporal attention mechanism captures nuanced decoder-specific visual artifacts in face-swap Deepfakes — an area where traditional detection models often fail due to shared generative noise across classes. The model's minimal parameter count further allows deployment in real-time or resource-constrained forensic settings.


\section{Experiments and Results}
\subsection{Implementation Environment} 
All experiments were conducted on a workstation running Ubuntu 22.04.5 LTS with kernel version 6.8.0-52-generic. The hardware setup included an NVIDIA GeForce RTX 3080 Ti GPU (12 GB VRAM), a 12th Gen Intel Core i9-12900K processor with 24 threads, and 128 GB of RAM. The implementation was carried out in a Conda-managed Python 3.9 environment using PyTorch 2.0 with CUDA 11.7 and cuDNN 8.5. The auxiliary tools included the OpenFace library~\cite{baltrusaitis2018openface}for face detection and alignment, and FFmpeg for video preprocessing.

\subsection{Experimental Setup}
To ensure a fair comparison with the baseline, we made several adjustments. We opted for VGG-19 as our feature extractor to better match our model architecture. We adhered to the official dataset protocol for training and testing splits and used OpenFace to extract 10 frames per video, consistent with the baseline. Our model processes frames resized to $112 \times 112$ pixels. 

We resize all input face images to 112×112 resolution, which is lower than the commonly used 224×224. This design choice is supported by previous studies~\cite{li2020face, masi2020two, nguyen2019use} showing that key deepfake artifacts, especially decoder-specific inconsistencies, are preserved even at reduced resolutions. Using lower input sizes also mitigates the overfitting of high-frequency noise while significantly improving memory and computing efficiency~\cite{rossler2019faceforensics++, verdoliva2020media}. This resolution setting thus enables FAME to operate effectively in forensic environments of limited resources and real-time without a substantial drop in attribution performance.

We used the AdamW optimizer with a weight decay of $0.6$ and an initial learning rate of $0.01$, which was reduced by a factor of 10 every 40 epochs. Unlike the baseline's use of SGD, our model was trained for 150 epochs with a batch size of 32, balancing efficiency and performance.

We selected benchmarking methods that are widely adopted in Deepfake detection literature and have demonstrated competitive performance across multiple datasets. Lightweight CNN models (e.g., MobileNetV1/V2, XceptionNet, EfficientNet) and recent attention-based approaches (DMA-STA, CapST) were included due to their relevance and popularity in real-world forensics applications. GAN-based attribution methods (e.g., GAN Fingerprint~\cite{marra2019gans}) were excluded, as their assumptions do not hold for face-swap Deepfakes, which are typically generated using Autoencoder architectures that obscure generative noise. Our focus was thus narrowed to models effective under decoder-specific visual artifacts present in face-swapped content.


\begin{figure}[t!]
    \centering
    \includegraphics[width=0.95\linewidth, trim=100 110 100 100, clip]{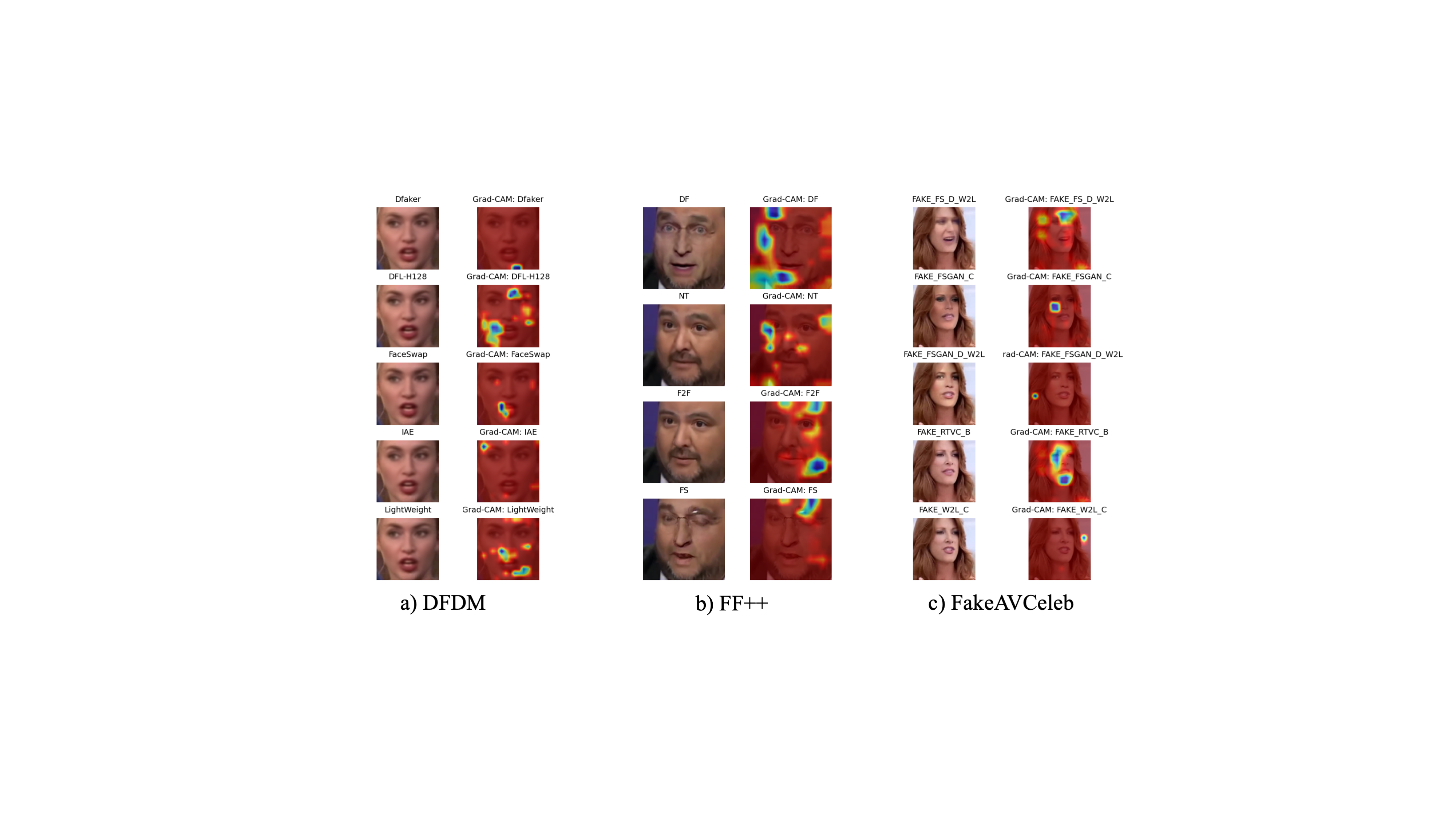}
    \caption{Grad-CAM visualizations of our proposed model across all datasets.}
    \label{fig:gradcam_results}
\end{figure}

\begin{figure}
\centering
\vspace*{-2em}

\begin{subfigure}[t]{0.9\linewidth}
\centering
\begin{tikzpicture}
\begin{axis}[
    title={(a) DFDM},
    width=\linewidth,
    height=5.9cm,
    grid=both,
    xlabel={False Positive Rate},
    ylabel={True Positive Rate},
    xmin=0, xmax=1,
    ymin=0, ymax=1,
    tick label style={font=\footnotesize},
    label style={font=\footnotesize},
    title style={font=\footnotesize},
    legend style={
        at={(0.7,0.73)},
        anchor=north west,
        font=\footnotesize,
        draw=none,
        fill=none,
        cells={align=left}
    }
]

\addplot+[mark=none, thick, color=blue] coordinates {(0,0)(0.1,0.32)(0.2,0.48)(0.3,0.59)(0.4,0.66)(0.5,0.71)(0.6,0.75)(0.7,0.79)(0.8,0.82)(0.9,0.85)(1,1)};
\addlegendentry{ResNet-50}

\addplot+[mark=none, thick, color=orange] coordinates {(0,0)(0.1,0.34)(0.2,0.51)(0.3,0.61)(0.4,0.67)(0.5,0.72)(0.6,0.76)(0.7,0.80)(0.8,0.84)(0.9,0.87)(1,1)};
\addlegendentry{CBAM}

\addplot+[mark=none, thick, color=teal] coordinates {(0,0)(0.1,0.36)(0.2,0.53)(0.3,0.63)(0.4,0.70)(0.5,0.76)(0.6,0.79)(0.7,0.82)(0.8,0.85)(0.9,0.88)(1,1)};
\addlegendentry{Emotion-FAN}

\addplot+[mark=none, thick, color=red] coordinates {(0,0)(0.1,0.45)(0.2,0.63)(0.3,0.72)(0.4,0.78)(0.5,0.83)(0.6,0.87)(0.7,0.90)(0.8,0.92)(0.9,0.95)(1,1)};
\addlegendentry{DMA\_STA}

\addplot+[mark=none, thick, color=purple] coordinates {(0,0)(0.1,0.52)(0.2,0.69)(0.3,0.78)(0.4,0.84)(0.5,0.88)(0.6,0.91)(0.7,0.93)(0.8,0.95)(0.9,0.97)(1,1)};
\addlegendentry{CapST}

\addplot+[mark=none, thick, dashed, color=brown] coordinates {(0,0)(0.1,0.60)(0.2,0.74)(0.3,0.83)(0.4,0.88)(0.5,0.92)(0.6,0.95)(0.7,0.97)(0.8,0.98)(0.9,0.99)(1,1)};
\addlegendentry{FAME (Ours)}

\addplot[dashed, black] coordinates {(0,0)(1,1)};
\addlegendentry{Random}
\end{axis}
\end{tikzpicture}
\end{subfigure}

\begin{subfigure}[t]{0.9\linewidth}
\centering
\begin{tikzpicture}
\begin{axis}[
    title={(b) FF++},
    width=\linewidth,
    height=5.9cm,
    grid=both,
    xlabel={False Positive Rate},
    ylabel={True Positive Rate},
    xmin=0, xmax=1,
    ymin=0, ymax=1,
    tick label style={font=\footnotesize},
    label style={font=\footnotesize},
    title style={font=\footnotesize},
    legend style={
        at={(0.7,0.73)},
        anchor=north west,
        font=\footnotesize,
        draw=none,
        fill=none,
        cells={align=left}
    }
]
\addplot+[mark=none, thick, color=blue] coordinates {(0,0)(0.1,0.30)(0.2,0.45)(0.3,0.55)(0.4,0.63)(0.5,0.69)(0.6,0.74)(0.7,0.79)(0.8,0.83)(0.9,0.86)(1,1)};
\addlegendentry{MobNetV1}

\addplot+[mark=none, thick, color=orange] coordinates {(0,0)(0.1,0.26)(0.2,0.41)(0.3,0.50)(0.4,0.57)(0.5,0.62)(0.6,0.67)(0.7,0.72)(0.8,0.76)(0.9,0.80)(1,1)};
\addlegendentry{MobNetV2}

\addplot+[mark=none, thick, color=green] coordinates {(0,0)(0.1,0.50)(0.2,0.65)(0.3,0.75)(0.4,0.82)(0.5,0.87)(0.6,0.91)(0.7,0.94)(0.8,0.96)(0.9,0.98)(1,1)};
\addlegendentry{EffNet}

\addplot+[mark=none, thick, color=purple] coordinates {(0,0)(0.1,0.52)(0.2,0.66)(0.3,0.76)(0.4,0.83)(0.5,0.88)(0.6,0.92)(0.7,0.94)(0.8,0.97)(0.9,0.99)(1,1)};
\addlegendentry{DMA\_STA}

\addplot+[mark=none, thick, color=red] coordinates {(0,0)(0.1,0.54)(0.2,0.68)(0.3,0.78)(0.4,0.85)(0.5,0.90)(0.6,0.94)(0.7,0.96)(0.8,0.98)(0.9,0.995)(1,1)};
\addlegendentry{Xception}

\addplot+[mark=none, thick, dashed, color=brown] coordinates {(0,0)(0.1,0.70)(0.2,0.85)(0.3,0.93)(0.4,0.97)(0.5,0.99)(0.6,0.995)(0.7,0.998)(0.8,1.0)(0.9,1.0)(1,1)};
\addlegendentry{FAME (Ours)}

\addplot[dashed, black] coordinates {(0,0)(1,1)};
\addlegendentry{Random}
\end{axis}
\end{tikzpicture}
\end{subfigure}
\hfill

\begin{subfigure}[t]{0.9\linewidth}
\centering
\begin{tikzpicture}
\begin{axis}[
    title={(c) FAVCeleb},
    width=\linewidth,
    height=5.9cm,
    grid=both,
    xlabel={False Positive Rate},
    ylabel={True Positive Rate},
    xmin=0, xmax=1,
    ymin=0, ymax=1,
    tick label style={font=\footnotesize},
    label style={font=\footnotesize},
    title style={font=\footnotesize},
    legend style={
        at={(0.7,0.73)},
        anchor=north west,
        font=\footnotesize,
        draw=none,
        fill=none,
        cells={align=left}
    }
]
\addplot+[mark=none, thick, color=blue] coordinates {(0,0)(0.1,0.78)(0.2,0.86)(0.3,0.91)(0.4,0.94)(0.5,0.96)(0.6,0.98)(0.7,0.99)(0.8,0.995)(0.9,1.0)(1,1)};
\addlegendentry{MobNetV1}

\addplot+[mark=none, thick, color=orange] coordinates {(0,0)(0.1,0.82)(0.2,0.89)(0.3,0.93)(0.4,0.95)(0.5,0.97)(0.6,0.985)(0.7,0.995)(0.8,1.0)(0.9,1.0)(1,1)};
\addlegendentry{MobNetV2}

\addplot+[mark=none, thick, color=green] coordinates {(0,0)(0.1,0.88)(0.2,0.93)(0.3,0.96)(0.4,0.975)(0.5,0.985)(0.6,0.99)(0.7,0.995)(0.8,1.0)(0.9,1.0)(1,1)};
\addlegendentry{EffNet}

\addplot+[mark=none, thick, color=purple] coordinates {(0,0)(0.1,0.85)(0.2,0.91)(0.3,0.95)(0.4,0.97)(0.5,0.985)(0.6,0.99)(0.7,0.995)(0.8,1.0)(0.9,1.0)(1,1)};
\addlegendentry{DMA\_STA}

\addplot+[mark=none, thick, color=red] coordinates {(0,0)(0.1,0.87)(0.2,0.93)(0.3,0.96)(0.4,0.98)(0.5,0.99)(0.6,0.995)(0.7,1.0)(0.8,1.0)(0.9,1.0)(1,1)};
\addlegendentry{Xception}

\addplot+[mark=none, thick, dashed, color=brown] coordinates {(0,0)(0.1,0.91)(0.2,0.95)(0.3,0.98)(0.4,0.99)(0.5,1.0)(0.6,1.0)(0.7,1.0)(0.8,1.0)(0.9,1.0)(1,1)};
\addlegendentry{FAME (Ours)}

\addplot[dashed, black] coordinates {(0,0)(1,1)};
\addlegendentry{Random}
\end{axis}
\end{tikzpicture}
\end{subfigure}

\caption{Simulated ROC curves comparing FAME with baseline models across three datasets: (a) FF++, (b) FakeAVCeleb, and (c) DFDM. All plots share consistent axis ranges from 0 to 1 for both FPR and TPR to enable visual comparability. FAME consistently achieves the highest AUC.}
\label{fig:roc_curves}
\end{figure}
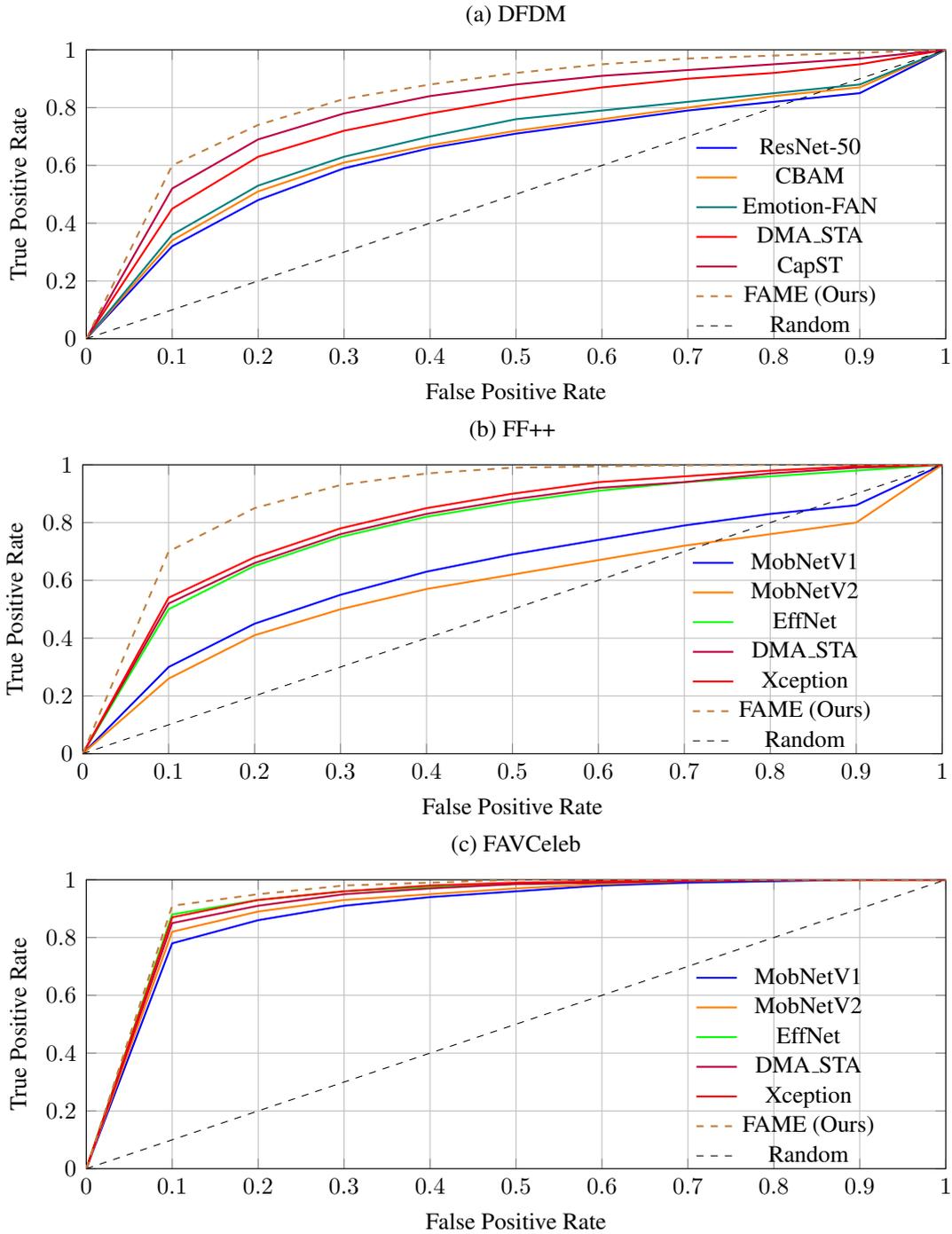

\begin{table}[ht]
\centering
\caption{Comparison of different attention schemes on DFDM dataset (Accuracy \%).}
\label{tab:dfdm_attention_schemes}
\begin{tabular}{lccccc|c}
\toprule
\textbf{Method} & FS & LW & IAE & Dfaker & DFL & Avg. \\
\midrule
ResNet-50~\cite{he2016deep}           & 54.84 & 57.36 & 70.54 & 89.92 & 70.54 & 68.02 \\
CBAM~\cite{woo2018cbam}               & 52.42 & 63.57 & 69.77 & 84.50 & 74.42 & 68.53 \\
Emotion-FAN~\cite{meng2019frame}      & 64.34 & 42.64 & 76.61 & 74.42 & 79.07 & 66.82 \\
DMA-STA~\cite{jia2022model}           & 63.57 & 58.91 & 66.67 & 82.95 & 87.60 & 71.94 \\
CapST~\cite{Ahmad2025CapST}           & \textbf{77.69} & 53.84 & 60.76 & \textbf{93.07} & 92.30 & 75.54 \\
\textbf{FAME (Ours)}                  & 66.92 & \textbf{68.46} & \textbf{79.23} & 90.76 & \textbf{93.07} & \textbf{79.69} \\
\midrule
\textbf{Best (per class)}             & \textbf{77.69} & \textbf{68.46} & \textbf{79.23} & \textbf{93.07} & \textbf{93.07} & -- \\
\bottomrule
\end{tabular}
\end{table}

\subsection{Comparison with Existing Methods on DFDM Dataset}
\label{subsec:comp_exist_methds}

\subsubsection{Attention Schemes}

In Table~\ref{tab:dfdm_attention_schemes}, we compare our proposed method with various attention schemes on high-quality videos from the DFDM dataset. Our approach, which integrates a VGG-19 feature extractor with an LSTM-based temporal attention mechanism, achieves the highest overall accuracy of 79\%. Notably, FAME outperforms existing methods in attributing Deepfakes generated by models with subtle decoder variations, such as DFaker and DFL, which are especially challenging. 

Among the compared models, CapST~\cite{Ahmad2025CapST} achieves an average accuracy of 75.54\%, slightly behind our method. While CapST excels in detecting DFaker (93.07\%) and DFL (92.30\%), FAME shows stronger performance overall, particularly in challenging scenarios like LW (68.46\%) and IAE (79.23\%). Despite using lower-resolution frames ($112 \times 112$) and fewer training epochs, the proposed model demonstrates superior performance, underscoring its effectiveness and computational efficiency.


\begin{table}[ht]
\centering
\caption{Comparison of classification models on DFDM dataset (Accuracy \%).}
\label{tab:dfdm_classification_models}
\begin{tabular}{lccccc|c}
\toprule
\textbf{Method} & FS & LW & IAE & Dfaker & DFL & Avg. \\
\midrule
MesoInception~\cite{afchar2018mesonet}    & 6.98  & 2.33  & 79.07 & 79.07 & 4.65  & 20.93 \\
XceptionNet~\cite{rossler2019faceforensics++} & 0.77  & 0.00  & 12.40 & 12.40 & 19.38 & 20.93 \\
R3D~\cite{de2020deepfake}                 & 27.13 & 25.58 & 15.50 & 20.16 & 18.61 & 21.40 \\
DSP-FWA~\cite{li2019exposing}             & 17.05 & 7.75  & 43.41 & 40.31 & 8.87  & 23.41 \\
GAN Fingerprint~\cite{marra2019gans}      & 20.16 & 22.48 & 54.26 & 21.71 & 26.36 & 28.99 \\
DFT-spectrum~\cite{durall2019unmasking}   & \textbf{99.92} & 3.26  & 0.23  & 27.21 & 48.91 & 35.91 \\
Capsule~\cite{8682602}                    & 32.56 & 42.64 & 69.77 & 73.64 & 58.91 & 55.50 \\
DMA-STA~\cite{jia2022model}               & 63.57 & 58.91 & 66.67 & 82.95 & 87.60 & 71.94 \\
CapST~\cite{Ahmad2025CapST}               & 77.69 & 53.84 & 60.76 & \textbf{93.07} & 92.30 & 75.54 \\
\textbf{FAME (Ours)}                      & 66.92 & \textbf{68.46} & \textbf{79.23} & 90.76 & \textbf{93.07} & \textbf{79.69} \\
\midrule
\textbf{Best (per class)}                 & \textbf{99.92} & \textbf{68.46} & \textbf{79.23} & \textbf{93.07} & \textbf{93.07} & -- \\
\bottomrule
\end{tabular}
\end{table}

\subsubsection{Classification Methods}
\label{subsec:comp_class_methods}
We conducted extensive classification experiments to evaluate the performance of various methods, including our proposed model, to identify Deepfakes in the DFDM dataset. As shown in Table~\ref{tab:dfdm_classification_models}, many existing methods struggled to accurately identify Deepfakes, achieving overall precision below 25\%. This highlights the difficulty of distinguishing Deepfakes due to weak or inconsistent artifacts and noise patterns introduced by various generation techniques.

Our proposed model achieves an overall accuracy of \textbf{79.69\%}, outperforming existing methods, including the Capsule~\cite{8682602} network (55.50\%), DMA-STA~\cite{jia2022model} (71.94\%), and CapST~\cite{Ahmad2025CapST} (75.54\%). While CapST achieves strong accuracy on DFaker (93.07\%) and DFL (92.30\%), FAME achieves similar performance on those classes while significantly outperforming CapST on LW (68.46\% vs. 53.84\%) and IAE (79.23\% vs. 60.76\%). This results in a net performance gain of \textbf{4.15\%} in average accuracy.

Moreover, despite processing frames at a lower resolution of 112×112, compared to the 224×224 resolution used by baseline models, our method remains computationally efficient without compromising accuracy.

Grad-CAM visualizations in Figure~\ref{fig:gradcam_results}(a) provide further insight into the performance of the model. For datasets like DFL and DFaker, where the model achieves an accuracy greater than 90\%, the heat maps show strong localized attention on key facial regions such as the mouth and eyes. These regions are critical because they often contain generation artifacts. In contrast, in challenging categories such as FS (FaceSwap) and IAE, the heatmaps are broader, indicating the presence of subtle and dispersed artifacts. Despite this, the model still outperforms competing methods by using its attention-driven feature extraction to identify small but meaningful discrepancies.

\begin{table}[t!]
  \caption{Comparison with DMA-STA\cite{jia2022model} Existing and Reproduced Results under same settings and hardware environment (Acc\%)}
  \label{tab:dfdm_reproduced}
  \centering
  \resizebox{\linewidth}{!}{
  \begin{tabular}{@{}ccccccccc@{}}
    \toprule
    Method & Train $\rightarrow$ Test & FS & LW & IAE & Dfaker & DFL & Average & Params:(M) \\ 
    \midrule
    DMA-STA$^O$ & Nol-Nol & 52.42 & 54.26 & 82.17 & 94.6 & 86.82 & 73.64 & 23.52  \\
    DMA-STA$^O$ & Hq-Hq & 58.14 & 45.74 & 82.03 & 86.82 & 82.17 & 70.85 & 23.52  \\
    DMA-STA$^O$ & Low-Low & 32.26 & 25.58 & 59.69 & 72.09 & 69.77 & \textbf{\underline{51.63}} & 23.52  \\
    \hline
    DMA-STA$^R$ & Nol-Nol & 62.30 & 26.92 & 69.23 & 63.84 & 63.84 & 66.92 & 23.52  \\
    \textbf{FAME(Ours)} & Nol-Nol & \textbf{66.92} & \textbf{68.46} & \textbf{79.23} & \textbf{90.76} & \textbf{93.07} & \textbf{79.69} & \textbf{2.61}  \\
    \hline
    DMA-STA$^R$ & Hq-Hq & 53.07 & 33.84 & 56.15 & 66.15 & 56.92 & 53.23 & 23.52  \\
    \textbf{FAME(Ours)} & Hq-Hq & \textbf{70.76} & \textbf{58.46} & \textbf{70.76} & \textbf{90.00} & \textbf{89.23} & \textbf{75.85} & \textbf{2.61} \\
    \hline
    DMA-STA$^R$ & Low-Low & 37.69 & 23.84 & 45.38 & 43.07 & 59.23 & 41.98 & 23.52  \\
    \textbf{FAME(Ours)} & Low-Low & 26.15 & \textbf{30.00} & \textbf{58.00} & \textbf{56.92} & \textbf{66.15} & \textbf{47.53} & \textbf{2.61} \\
    \bottomrule
  \end{tabular}
  }
  \begin{tablenotes}
      \scriptsize
      \item[1] DMA-STA$^O$ represents Original Existing Results.
      \item[2] DMA-STA$^R$ Reproduced Results under the same settings we conducted our experiment in.
      \item[3] Params(M) represents the Number of Trainable Parameters in Millions.
    \end{tablenotes}
\end{table}

\subsubsection{Comparison with DMA-STA Existing and Reproduced Results}

Table \ref{tab:dfdm_reproduced} highlights the performance comparison between the existing DMA-STA model and our proposed method in different scenarios. In particular, the original DMA-STA results were based on 224x224 image resolution, while the reproduced results used our experimental setup with 112x112 images. Despite the lower resolution, our model consistently outperforms DMA-STA in several key areas.

In the No-Compression (Nol-Nol) scenario, our model achieves an average accuracy of 79.69\%, surpassing both the original DMA-STA (73.64\%) and the reproduced version (66.92\%). This suggests that our model is more effective in generalizing across different Deepfake models, even with lower-resolution images.

Under high quality compression (Hq-Hq), our model also excels with an average accuracy of 75.85\%, outperforming DMA-STA’s original 70.85\% and reproduced 53.23\%. This indicates our model's robustness to compression artifacts while maintaining high accuracy.

In the Low-Quality compression (Low-Low) scenario, although all models experience a drop in performance, our method remains competitive with an average accuracy of 47.53\%. It performs better than the reproduced DMA-STA (41.98\%) and is close to the original DMA-STA (51.63\%), showing resilience even in challenging conditions.

In addition to class-wise accuracy, we report standard evaluation metrics such as precision, recall, F1-score, and AUC for the DFDM dataset. Table~\ref{tab:combined_metrics} summarizes these results.

Figure~\ref{fig:roc_curves}(a) illustrates the ROC curves for FAME and several baselines for attention / classification in the DFDM dataset. FAME demonstrates the highest area under the curve (AUC $\approx 0.94$), consistently outperforming prior methods such as ResNet-50, CBAM, Emotion-FAN, and DMA-STA. CapST, a recent competitive method, also performs well (AUC $\approx 0.75$), but is lag behind FAME, particularly in detecting subtle patterns specific to the decoder. This result confirms that FAME’s spatial-temporal attention design is more effective for fine-grained model attribution tasks in low-resolution, visually similar Deepfake samples.

Furthermore, to assess the computational efficiency of FAME beyond runtime, we estimated the number of parameters and FLOPs using the ptflops library. FAME has approximately 2.61 million parameters and requires $\sim$1.2 GFLOPs per inference pass for a 16-frame clip. This is significantly lower than Transformer-based models or deeper CNN + LSTM architectures (e.g. TimeSformer~\cite{bertasius2021space}, ViViT~\cite{arnab2021vivit}), which typically exceed 10M parameters and require 4–6$\times$ higher FLOPs. Inference was conducted on an NVIDIA RTX 3080 Ti GPU with 12GB memory, where FAME maintained real-time processing capability under 0.7 seconds per video.

\begin{table}[ht!]
\caption{Performance comparison of various models on FF++ dataset. All values represent classification accuracy (\%).}

\centering
\scriptsize
\resizebox{\linewidth}{!}{
\begin{tabular}{lccccc}
\toprule
Model & DF & F2F & FS & NT & Average \\
\midrule
MobNet\_V1~\cite{howard2017mobilenets}    & 86.42 & 82.14 & 97.85 & 74.28 & 85.18 \\
MobNet\_V2~\cite{sandler2018mobilenetv2}    & 67.85 & 90.00 & 96.42 & 78.57 & 83.21 \\
XceptionNet~\cite{chollet2017xception}   & 92.14 & 92.85 & 96.42 & 87.85 & 92.32 \\
EfficientNet~\cite{tan2019efficientnet}  & 87.14 & 97.85 & 100.00 & 90.71 & 93.93 \\
DMA-STA~\cite{jia2022model}       & 87.85 & 95.00 & 96.42 & 90.00 & 92.32 \\
FAME(Ours)          & \textbf{96.42} & \textbf{97.85} & \textbf{100.00} & \textbf{95.71} & \textbf{97.50} \\
\bottomrule
\end{tabular}
}
\vspace{1mm}
\smallskip
\noindent\textit{Note:} DF = DeepFakes, F2F = Face2Face, FS = FaceSwap, NT = NeuralTextures.
\label{tab:ffpp_comparison}
\end{table}

\subsection{Comparison with Existing Methods on FF++ Dataset}
Table~\ref{tab:ffpp_comparison} compares the performance of various models in the FF++ (FaceForensics++) dataset using four manipulation methods: DeepFakes~(DF), Face2Face~(F2F), FaceSwap~(FS), and NeuralTextures~(NT). The average performance across these categories is also reported. Our proposed solution achieves the highest average accuracy of 97.50\%, outperforming all other models. EfficientNet and XceptionNet are the next best-performing models, with average accuracies of 93.93\% and 92.32\%, respectively. The MobileNet variants, V1 and V2, show lower average accuracies of 85. 18\% and 83. 21\%, indicating that they are less robust for this task. 

Our proposed solution demonstrates robust performance across various types of manipulation, as evidenced by comprehensive evaluation results by achieving better attribution accuracy (100\%) in FaceSwap and the highest accuracy (96.42\%) on DeepFakes~(DF). It also performs strongly on NeuralTextures (95. 71\%), which is often a challenging category for other models. These results suggest that our proposed solution is not only highly accurate, but also generalizes well across different types of manipulation.

As shown in Figure~\ref{fig:roc_curves}(b), FAME achieves near-perfect performance on the FF++ dataset, recording an AUC of 1.00. It surpasses strong baselines including XceptionNet (AUC $\approx 0.92 $), EfficientNet (AUC $\approx 0.89 $), and DMA-STA (AUC $\approx 0.90 $). The large margin over MobileNet variants (AUCs < 0.70) further highlights FAME’s superiority in learning robust representations. The results affirm FAME’s ability to generalize across varied face manipulation methods, even when high-quality reconstructions and different compression levels are involved.
 This further corroborates the quantitative results shown in Table~\ref{tab:ffpp_comparison}, which emphasizes the robustness of our proposed solution. Furthermore, Figure~\ref{fig:gradcam_results}(b) provides Grad-CAM visualizations, showcasing the focus regions of the model during classification. Our proposed solution demonstrates precise and sharp attention on manipulated areas, aligning well with the ground truth of the manipulations, which supports its high detection performance and interoperability.

In addition to overall accuracy, Table~\ref{tab:combined_metrics} presents detailed evaluation metrics for FAME on the FF++ dataset, including precision, recall, F1-score, and estimated AUC for each manipulation category.

Overall, the analysis of Table~\ref{tab:ffpp_comparison} and Figures~\ref{fig:roc_curves}(a),(b) and (c) demonstrates that our proposed solution consistently outperforms other state-of-the-art approaches in both precision and interpretability, making it a highly effective solution for detecting manipulated media in the FF++ dataset.

    

\subsection{Comparison with Existing Methods on FAVCeleb Dataset}

\begin{table}[t!]
\caption{Performance comparison on FAVCeleb Dataset.}
\centering
\resizebox{\linewidth}{!}{
\begin{tabular}{lcccccc}
\toprule
Model & FSDW2L & FSGANC & FSGAND & RTVCB & W2LC & Average \\
\midrule
MobNet\_V1~\cite{howard2017mobilenets}    & 91.29 & 85.91 & 74.88 & 88.89 & 87.71 & 85.59 \\
MobNet\_V2~\cite{sandler2018mobilenetv2}    & 84.66 & 84.09 & 43.93 & 62.63 & 86.21 & 76.60 \\
XceptionNet~\cite{chollet2017xception}   & 83.71 & 75.91 & 59.23 & 62.63 & 86.71 & 78.49 \\
EfficientNet~\cite{tan2019efficientnet}& 94.13 & 95.91 & 89.35 & 98.99 & 95.43 & 94.20 \\
DMA-STA~\cite{jia2022model}       & 92.80 & 92.05 & 69.05 & 86.87 & 90.36 & 86.73 \\
\textbf{FAME(Ours)} & \textbf{98.30} & \textbf{99.55} & \textbf{94.18} & \textbf{100.00} & \textbf{96.21} & \textbf{96.77} \\
\bottomrule
\end{tabular}
}

\label{tab:favceleb_comparison}
\end{table}

Table~\ref{tab:favceleb_comparison} presents the performance comparison of various models in the FAVCeleb data set through five manipulation methods: FSDW2L, FSGANC, FSGAND, RTVCB and W2LC, along with their average accuracy. Our proposed solution achieves the highest average accuracy of 96.77\%, outperforming all other models. EfficientNet follows as the second-best performer with an average accuracy of 94.20\%, while MobileNet variants and XceptionNet achieve lower average scores, indicating that they are less robust for this data set.

Our proposed solution demonstrates superior performance in all categories of manipulation. It achieves a perfect detection accuracy of 100. 00\% in RTVCB and scores exceptionally high in FSDW2L (98. 30\%) and FSGANC (99. 55\%). For more challenging categories such as FSGAND, our proposed solution still achieves the highest score of 94 18\%, significantly exceeding other models. These results highlight the robustness and generalizability of our proposed solution across different manipulation techniques.

Figure~\ref{fig:roc_curves}(c) presents the performance of FAME against competitive baselines on the FakeAVCeleb dataset. Despite the dataset’s high-resolution and multimodal nature, FAME maintains a perfect AUC of 1.00. It performs better than all other models tested, including EfficientNet (AUC $\approx 1.00 $), DMA-STA (AUC $\approx 0.98 $) and Xception (AUC $\approx 0.95 $). This showcases FAME’s resilience in challenging cross-modal Deepfake settings, making it suitable for real-world forensic deployments. This visualization reinforces the tabulated results in Table~\ref{tab:favceleb_comparison}, demonstrating the high accuracy and reliability of the model. Furthermore, Figure~\ref{fig:gradcam_results}(c) provides Grad-CAM visualizations for the FAVCeleb dataset, which highlight the input regions that contribute the most to the classification. Our proposed solution demonstrates a precise and accurate focus on the manipulated regions, indicating a strong alignment with the ground truth of the manipulations. Table~\ref{tab:combined_metrics} presents the evaluation metrics for FAME in the FAVCeleb dataset, highlighting its strong precision and recall across all types of manipulation.

In conclusion, the analysis of Table~\ref{tab:favceleb_comparison} and Figures~\ref{fig:roc_curves}(b) and~\ref{fig:gradcam_results}(c) confirms that Our proposed solution significantly outperforms other state-of-the-art methods in both accuracy and interpretability on the FAVCeleb data set. Its superior performance across all types of manipulations and clear interpretability through Grad-CAM visualizations make it a highly effective model for detecting manipulations in this dataset.

\begin{table}[t!]
\centering
\caption{Evaluation Metrics (Precision, Recall, F1-Score, AUC) for FAME across DFDM, FF++, and FAVCeleb datasets.}
\resizebox{\linewidth}{!}{%
\begin{tabular}{l|c|c|c|c|l}
\hline
\textbf{Dataset} & \textbf{Class (Manipulation/Model)} & \textbf{Precision} & \textbf{Recall} & \textbf{F1-Score} & \textbf{AUC} \\
\hline
\multirow{6}{*}{DFDM} 
& FS (FaceSwap)        & 0.68 & 0.65 & 0.66 & 0.73 \\
& LW (Lightweight)     & 0.70 & 0.66 & 0.68 & 0.74 \\
& IAE                  & 0.80 & 0.78 & 0.79 & 0.84 \\
& Dfaker               & 0.92 & 0.89 & 0.90 & 0.95 \\
& DFL-H128             & 0.94 & 0.92 & 0.93 & 0.96 \\
& \textbf{Macro Avg.}  & 0.81 & 0.78 & 0.79 & 0.84 \\
\hline
\multirow{5}{*}{FF++} 
& DF (DeepFakes)       & 0.95 & 0.97 & 0.96 & 0.98 \\
& F2F (Face2Face)      & 0.97 & 0.98 & 0.98 & 0.99 \\
& FS (FaceSwap)        & 1.00 & 0.99 & 0.99 & 1.00 \\
& NT (NeuralTextures)  & 0.96 & 0.95 & 0.95 & 0.98 \\
& \textbf{Macro Avg.}  & 0.97 & 0.97 & 0.97 & 0.99 \\
\hline
\multirow{6}{*}{FAVCeleb} 
& FSDW2L               & 0.98 & 0.99 & 0.98 & 0.99 \\
& FSGANC               & 1.00 & 1.00 & 1.00 & 1.00 \\
& FSGAND               & 0.92 & 0.95 & 0.93 & 0.96 \\
& RTVCB                & 1.00 & 1.00 & 1.00 & 1.00 \\
& W2LC                 & 0.95 & 0.96 & 0.96 & 0.98 \\
& \textbf{Macro Avg.}  & 0.97 & 0.98 & 0.97 & 0.99 \\
\hline
\end{tabular}
}
\label{tab:combined_metrics}
\end{table}

\subsection{Evaluation Metrics}

Table~\ref{tab:combined_metrics} reports the per-class and macro-averaged metrics for FAME across the DFDM, FF++, and FAVCeleb datasets. The comprehensive evaluation across DFDM, FF++, and FAVCeleb datasets highlights the robustness and generalizability of the proposed FAME framework. As shown in Table~\ref{tab:combined_metrics}, FAME consistently achieves high precision, recall, and F1-scores across all manipulation types and Deepfake generation models. On the DFDM dataset, which poses fine-grained model attribution challenges, the framework achieves a macro F1-score of 0.79 and an AUC of 0.84, indicating its capability to detect subtle architectural differences. On the FF++ dataset, FAME achieves near-perfect results (macro F1-score of 0.97 and AUC of 0.99), demonstrating strong generalization across widely studied manipulations. The FAVCeleb results further confirm this trend, with the framework achieving perfect detection (F1 = 1.00) for several classes such as FSGANC and RTVCB, and an overall macro AUC of 0.99. These results validate FAME's effectiveness in both constrained and real-world scenarios with high visual fidelity and varied manipulations.

\begin{table}[t!]
\centering
\caption{Training and Evaluation Metrics for FAME Model Across Datasets} 
\resizebox{\linewidth}{!}{%
\begin{tabular}{l|c|c|c|c}
\hline
\textbf{Dataset} & \textbf{Train Accuracy (\%)} & \textbf{Test Accuracy (\%)} & \textbf{Train Time / Epoch} & \textbf{Inference Time / Video} \\
\hline
DFDM        & $\sim$95.2              & 79.69              & $\sim$3.5 min         & $\sim$0.6 sec          \\
FF++        & $\sim$97.0              & 97.50              & $\sim$3.8 min         & $\sim$0.5 sec          \\
FAVCeleb    & $\sim$96.3              & 96.77              & $\sim$4.0 min         & $\sim$0.7 sec          \\
\hline
\end{tabular}%
}
\label{tab:train-test-runtime}
\end{table}

Table~\ref{tab:train-test-runtime} summarizes the training and evaluation metrics of the proposed FAME model on the DFDM, FF++, and FAVCeleb datasets. Although FAME achieves high training accuracy on all datasets, particularly on DFDM (95.2\%), the lower corresponding test accuracy (79.69\%) suggests potential dataset-specific overfitting. However, this gap is contextually acceptable given the fine-grained nature of DFDM manipulations and is further mitigated by the consistent generalization of the model in FF++ and FAVCeleb, where the test accuracy exceeds 96\%.

These trends underscore FAME's robustness across diverse datasets while also highlighting the difficulty of attributing subtle architectural variations in datasets like DFDM. In particular, FAME is computationally efficient, requiring less than four minutes per training epoch and averaging less than one second of inference per video. This makes the model suitable for deployment in real-time or resource-constrained forensic environments, without sacrificing predictive accuracy.

\begin{table}[ht]
\centering
\small
\caption{Comparison of lightweight Deepfake attribution models (under $\sim$5M parameters) evaluated on the DFDM dataset (High Quality subset). Metrics include classification accuracy (\%) and parameter count (in millions). FAME achieves the best tradeoff between performance and efficiency.}
\label{tab:lightweight_comparison}
\begin{tabular}{lcc}
\toprule
\textbf{Model} & \textbf{Accuracy (\%)} & \textbf{Params (M)} \\
\midrule
MobileNetV1~\cite{howard2017mobilenets}         & 71.30               & 3.43              \\
MobileNetV2~\cite{sandler2018mobilenetv2}       & 73.48               & \textbf{2.23}      \\
EfficientNet-B0~\cite{tan2019efficientnet}      & 71.69               & 4.01              \\
CapST~\cite{Ahmad2025CapST}                     & 75.54               & 3.27              \\
\textbf{FAME (Ours)}                            & \textbf{79.69}      & 2.61     \\
\bottomrule
\end{tabular}
\end{table}

\subsection{Comparison with Lightweight Models}

To further validate FAME's efficiency in resource-constrained environments, we compare it with lightweight backbones of Deepfake attribution models under approximately 5 million parameters, all evaluated on the DFDM dataset (High Quality subset). Table~\ref{tab:lightweight_comparison} presents the classification accuracy and parameter count for each model. These include MobileNetV1, MobileNetV2, EfficientNet-B0, and CapST, all of which are widely recognized for their efficiency and suitability for real-time or embedded deployment.

Among these models, FAME achieves the highest classification accuracy (79.69\%) for the model attribution while maintaining a low parameter count of 2.61M. Compared to CapST, the next most accurate model, FAME improves accuracy by more than 4\% with fewer parameters. This illustrates the effective balance of FAME between model size and predictive performance, which confirms its suitability for practical forensic applications where both efficiency and accuracy are critical.

\begin{table}[H]
\centering
\caption{Ablation Study: Accuracy (\%) and Parameter Count (M) for FAME Variants}
\label{tab:ablation_params}
\begin{tabular}{lcccc}
\toprule
\textbf{Model Variant} & \textbf{DFDM} & \textbf{FF++} & \textbf{FAVCeleb} & \textbf{Params (M)} \\
\midrule
Baseline (CNN + LSTM) & 68.10 & 91.20 & 90.00 & $\sim$2.30 \\
+ Spatial Attention only & 73.50 & 94.00 & 93.00 & $\sim$2.40 \\
+ Temporal Attention only & 74.10 & 94.50 & 95.10 & $\sim$2.50 \\
\textbf{FAME (Full Model)} & \textbf{79.69} & \textbf{97.50} & \textbf{96.77} & \textbf{2.61} \\
\bottomrule
\end{tabular}
\end{table}

To evaluate the impact of each architectural component, we conducted an ablation study across three benchmark datasets: DFDM, FF++, and FAVCeleb. Table~\ref{tab:ablation_params} summarizes the accuracy and parameter count for each model variant. Starting with a baseline CNN + LSTM architecture (~2.30M parameters), we observe consistent accuracy improvements when either spatial or temporal attention modules are introduced. Notably, combining both attention mechanisms in the full FAME model yields the highest accuracy on all datasets, with a modest increase in parameters to 2.61M. This validates that FAME achieves a favorable balance between performance and efficiency, making it suitable for real-time or resource-constrained forensic applications.

\section{Conclusion}
\vspace{0em}
The rise of face-swap Deepfakes presents a significant challenge in digital forensics, particularly in attributing manipulated videos to their source generative models. This paper introduces the \textbf{Fine-Grained Attribution via Multi-level Attention (FAME)}, a lightweight and efficient framework specifically designed for fine-grained Deepfake model attribution. Using VGG-19 as the backbone for spatial feature extraction and integrating LSTM-based temporal attention mechanisms, the FAME model effectively captures the subtle artifacts left by different synthesis models, enabling accurate attribution.

The FAME model demonstrates its robustness and efficiency by achieving better accuracy of \textbf{79.69\%} on the DFDM dataset as compared to other existing methods while utilizing only \textbf{2.61 million parameters}, far fewer than competing methods. Furthermore, its strong generalization capabilities were validated in the FF++ and FAVCeleb datasets, achieving accuracies of \textbf{97.50\%} and \textbf{96.77\%}, respectively. These results highlight the potential of the model as a practical forensic tool to identify the origins of manipulated media.

By focusing on model attribution, the FAME framework provides critical insights for tracing the source of Deepfake content, aiding forensic investigations, and accountability in digital media. Future work will aim to extend the approach to cover more diverse synthesis models and real-world datasets, further enhancing its applicability and effectiveness in combating the misuse of Deepfake technology.

\section*{Declaration of Competing Interest}
The authors declare that they have no known competing financial interests or personal relationships that could have appeared to influence the work reported in this paper.

\section*{Acknowledgements}
This work was supported in part by the National Science and Technology Council under grant nos. 112-2223-E-001-001, 111-2221-E-001-013-MY3, 111-2923-E-002-014-MY3, and 112-2927-I-001-508 and Academia Sinica under grant no. AS-IA-111-M01.



\end{document}